\RequirePackage[final]{graphicx}
\documentclass[sigconf,balance,natbib=false]{acmart}

\copyrightyear{2021} 
\acmYear{2021} 
\setcopyright{acmlicensed}
\acmConference[SEAMS'21]{IEEE/ACM 16th International Symposium on Software Engineering for Adaptive and Self-Managing Systems}{May 23--24, 2020}{Madrid, Spain}

\usepackage{hyperref}
\usepackage[plain]{fancyref}
\usepackage{ifdraft}

\usepackage[inline]{enumitem}
\usepackage{xcolor}
\usepackage{booktabs}
\usepackage{xspace}
\usepackage[final]{listings}
\usepackage{acronym}
\usepackage{url}

\usepackage{algorithmicx}
\usepackage{algpseudocode}
\usepackage{algorithm}
\usepackage{verbatim}

\usepackage{tabularx}

\usepackage{etoolbox}
\makeatletter
\patchcmd{\@makecaption}
  {\scshape}
  {}
  {}
  {}
\patchcmd{\@makecaption}
  {\\}
  {.\ }
  {}
  {}

\usepackage
  [backend=bibtex,
   style=trad-abbrv,
   maxnames=3,
   natbib=true,
   giveninits=true,
   hyperref=true,
   url=false,
   doi=false,
   isbn=false,
   defernumbers]{biblatex}
\AtEveryBibitem
{
   \clearlist{address}
   \clearfield{date}
   \clearfield{doi}
   \clearfield{eprint}
   \clearfield{isbn}
   \clearfield{issn}
   \clearfield{month}
   \clearfield{note}
   \clearlist{location}
   \clearfield{series}
   \clearfield{url}
   \clearname{editor}
   \ifentrytype{inproceedings}
     {\clearfield{day}
      \clearfield{month}
      \clearfield{volume}}{}
}

\DeclareFieldFormat*{title}{\textsl{#1}\isdot}
\DeclareFieldFormat*{journaltitle}{#1}
\DeclareFieldFormat*{booktitle}{#1}
\DeclareSourcemap
 {\maps[datatype=bibtex,overwrite]
   {
    \map {
      \step[fieldsource=booktitle,
       match=\regexp{[Pp]roceedings}, replace={Proc.}]
      \step[fieldsource=booktitle,
       match=\regexp{[Ii]nternational\s+[Cc]onference}, replace={Intl. Conf.}]
      \step[fieldsource=booktitle,
       match=\regexp{[Ii]nternational\s+[Ww]orshop}, replace={Intl. Workshop.}]
    }
    \map{
      \step[fieldset=journal,
       match=\regexp{[Jj]ournal}, replace={Jour.}]
	  \step[fieldset=journal,
       match=\regexp{[Tt]ransactions}, replace={Trans.}]
    }
    \map {
      \step[fieldsource=booktitle,
       match=\regexp{[Pp]roceedings\s+of\s+the.+[Ee]uropean\s+[Cc]onference\s+in}, replace={European Conf. in}]}
    \map{
      \step[fieldsource=booktitle,
       match=\regexp{In\s+[Pp]roceedings\s+of\s+the.+[Ss]ymposium\s+on}, replace={Symp. on}]}
    \map{
      \step[fieldsource=booktitle,
       match=\regexp{[Pp]roceedings\s+of\s+the.+[Ii]nternational\s+[Cc]onference\s+on}, replace={Intl. Conf. on}]}
    \map{
      \step[fieldsource=booktitle,
       match=\regexp{[Pp]roceedings\s+of\s+the.+[Ii]nternational\s+[Ww]orkshop\s+on}, replace={Intl. Workshop on}]}}}

\definecolor{OliveGreen}{rgb}{0,0.6,0.3}

\renewcommand{\lstlistingname}{Snippet}
\newcommand*{\fancyreflstlabelprefix}{lst}
\newcommand*{\Freflstname}{\lstlistingname}
\newcommand*{\freflstname}{\MakeLowercase{\lstlistingname}}
\Frefformat{vario}{\fancyreflstlabelprefix}%
  {\Freflstname\fancyrefdefaultspacing#1#3}
\frefformat{vario}{\fancyreflstlabelprefix}%
  {\freflstname\fancyrefdefaultspacing#1#3}
\Frefformat{plain}{\fancyreflstlabelprefix}%
  {\Freflstname\fancyrefdefaultspacing#1}
\frefformat{plain}{\fancyreflstlabelprefix}%
  {\freflstname\fancyrefdefaultspacing#1}

\newcommand{\algname}{Algorithm}
\newcommand*{\fancyrefalglabelprefix}{alg}
\newcommand*{\Frefalgname}{\algname}
\newcommand*{\frefalgname}{\MakeLowercase{\algname}}
\Frefformat{vario}{\fancyrefalglabelprefix}%
  {\Frefalgname\fancyrefdefaultspacing#1#3}
\frefformat{vario}{\fancyrefalglabelprefix}%
  {\frefalgname\fancyrefdefaultspacing#1#3}
\Frefformat{plain}{\fancyrefalglabelprefix}%
  {\Frefalgname\fancyrefdefaultspacing#1}
\frefformat{plain}{\fancyrefalglabelprefix}%
  {\frefalgname\fancyrefdefaultspacing#1}

\newcommand*{\fancyreflnlabelprefix}{ln}
\newcommand*{\Freflnname}{Line}
\newcommand*{\freflnname}{\MakeLowercase{\Freflnname}}
\Frefformat{vario}{\fancyreflnlabelprefix}%
  {\Freflnname\fancyrefdefaultspacing#1#3}
\frefformat{vario}{\fancyreflnlabelprefix}%
  {\freflnname\fancyrefdefaultspacing#1#3}
\Frefformat{plain}{\fancyreflnlabelprefix}%
  {\Freflnname\fancyrefdefaultspacing#1}
\frefformat{plain}{\fancyreflnlabelprefix}%
  {\freflnname\fancyrefdefaultspacing#1}

\lstdefinelanguage{JavaScript}{
keywords={typeof, new, true, false, catch, function, return, null, catch, switch, var, if, in, for, while, do, else, case, break, throw, this, instanceof},
keywordstyle=\color{purple}\bfseries,
ndkeywords={},
ndkeywordstyle=\color{blue}\bfseries,
identifierstyle=\color{black},
sensitive=false,
comment=[l]{//},
morecomment=[s]{/*}{*/},
commentstyle=\color{OliveGreen}\ttfamily,
stringstyle=\color{OliveGreen}\ttfamily,
morestring=[b]',
morestring=[b]"
}

\lstdefinestyle{floating}{%
  frame=none,
  float=htb,
  captionpos=b,
  aboveskip=0pt,
  belowskip=0pt
}

\lstdefinestyle{ctxtraits}
 {language=JavaScript,
  frame=lines,
  showstringspaces=false,
  keywordstyle=\tt\bf,
  tabsize=3,
  style=floating,
  morekeywords={Trait, cop, Context, activate, deactivate, adapt, addObjectPolicy, manager}
}

\lstnewenvironment{ctxtraits}[1][]
 {\lstset{style=ctxtraits,#1}}{}

\newcommand{\eg}{\emph{e.g.,}\xspace}
\newcommand{\ie}{\emph{i.e.,}\xspace}

\definecolor{author}{rgb}{.5, .5, .5}
\definecolor{comment}{rgb}{.1, .0, .9}
\definecolor{note}{rgb}{.9, .4, .0}
\definecolor{idea}{rgb}{.1, .7, .0}
\definecolor{missing}{rgb}{.9, .1, .0}

\algnewcommand{\LeftComment}[1]{\Statex \(\triangleright\) #1}

\acrodef{AST}{Abstract Syntax Tree}
\acrodef{BPEL}{Business Process Execution Language}
\acrodef{BPM}{Business Process Modeling}
\acrodef{CIA}[ComInA]{Composing Interacting Adaptations}
\acrodef{COP}{Context-oriented Programming}
\acrodef{IOT}[IoT]{Internet of Things}
\acrodef{MAS}{Multi-Agent System}
\acrodef{ML}{Machine Learning}
\acrodef{QOS}[QoS]{Quality of Service}
\acrodef{RL}{Reinforcement Learning}
\acrodef{SOC}{Service-oriented Computing}
\acrodef{SPL}{Software Product Line}
\acrodef{SUDS}{Sustainable Urban Drainage System}

\acused{BPEL}
\acused{QOS}

\let\orig@figure\figure
\renewcommand*{\figure}[1][]{\orig@figure[#1]\vspace{-2.2ex}} 
\let\orig@endfigure\endfigure
\renewcommand*{\endfigure}{\vspace{-1ex}\orig@endfigure} 
\let\orig@algorithm\algorithm
\renewcommand*{\algorithm}[1][]{\orig@algorithm[#1]\vspace{-0.2ex}} 
\let\orig@endalgorithm\endalgorithm
\renewcommand*{\endalgorithm}{\vspace{-0.7ex}\orig@endalgorithm} 

\usepackage{caption}
\usepackage{array}
\usepackage{makecell}
\newcolumntype{x}[1]{>{\centering\arraybackslash}p{#1}}

\usepackage{tikz}

\makeatother

\acrodef{AST}{Abstract Syntax Tree}
\acrodef{BPEL}{Business Process Execution Language}
\acrodef{BPM}{Business Process Modeling}
\acrodef{CIA}[ComInA]{Composing Interacting Adaptations}
\acrodef{COP}{Context-oriented Programming}
\acrodef{IOT}[IoT]{Internet of Things}
\acrodef{MAS}{Multi-Agent System}
\acrodef{ML}{Machine Learning}
\acrodef{QOS}[QoS]{Quality of Service}
\acrodef{RL}{Reinforcement Learning}
\acrodef{SOC}{Service-oriented Computing}
\acrodef{SPL}{Software Product Line}
\acrodef{SUDS}{Sustainable Urban Drainage System}

\acused{BPEL}
\acused{QOS}

\makeatletter
\hypersetup
 {pdfauthor={\@author}, 
  pdftitle={\@title},
  pdfsubject={.},
  pdfkeywords={}}
\makeatother

\begin{document}

\title{Adaptation to Unknown Situations as the Holy Grail of Learning-Based Self-Adaptive Systems: Research Directions}

\author{Ivana Dusparic}
\affiliation{School of Computer Science and Statistics\\
	Trinity College Dublin, Ireland\\
	ivana.dusparic@scss.tcd.ie}

\author{Nicol\'as Cardozo}
\affiliation{%
  Systems and Computing Engineering Department 
  Universidad de los Andes, Colombia\\
  n.cardozo@uniandes.edu.co
}




\renewcommand{\shortauthors}{I. Dusparic, N. Cardozo}


\begin{abstract}	
Self-adaptive systems continuously adapt to changes in their 
execution environment. Capturing all possible changes to define 
suitable behaviour beforehand is unfeasible, or even impossible in 
the case of unknown changes, hence human intervention may be required. 
We argue that adapting to unknown situations is the ultimate challenge 
for self-adaptive systems.
Learning-based approaches are used to learn the suitable 
behaviour to exhibit in the case of unknown situations, to minimize 
or fully remove human intervention. While such approaches can, to 
a certain extent, generalize existing adaptations to new situations, 
there is a number of breakthroughs that need to be achieved before 
systems can adapt to general unknown and unforeseen situations. 
We posit the research directions that need to be explored to achieve 
unanticipated adaptation from the perspective of learning-based 
self-adaptive systems. At minimum, systems need to define internal 
representations of previously unseen situations on-the-fly, 
extrapolate the relationship to the previously encountered 
situations to evolve existing adaptations, and 
reason about the feasibility of achieving their intrinsic goals in 
the new set of conditions. We close discussing whether, even when we 
can, we should indeed build systems that define their own behaviour 
and adapt their goals, without involving a human supervisor. 
\end{abstract}

%
%



\maketitle


\section{Adapting to Unknown Situations}
\label{sec:intro}

Self-adaptive systems are capable of adapting to changes in their 
execution environment in order to continue meeting their specified 
goals. Causes of adaptation can be internal or external, and different 
adaptation mechanisms might be suitable based on the type of 
adaptation trigger~\cite{deLemos2013}. One of the main challenges in 
self-adaptation is adapting to previously unknown situations. 
Approaches distinguish between so-called \emph{known unknowns}, which 
systems have not previously experienced but have mechanisms to reason about 
and react to (\eg by evolving, learning, or involving a human), and 
\emph{unknown unknowns}, which neither the system nor the designers 
of the system have foreseen~\cite{Weyns18}. 
Learning-based techniques are used to enable self-adaptive systems to 
adapt to known unknowns~\cite{dangelo19}. In this position paper we 
argue that adaptation to unknown unknowns is the ultimate challenge of 
self-adaptive systems and that learning-based self-adaptation is the key 
to achieving it. However, this  requires building and modifying 
learning processes at run time. Previous work has started to move 
towards dealing with unknown situations, however, the main challenges 
for adaptations to the unknown remain unaddressed. In the rest of 
this paper we outline the three main research directions required to 
achieve this goal.



\section{Learning-based Self-Adaptivity to Unknown Situations}
\label{sec:agenda}

While numerous issues in many aspects of self-adaptive system 
specification, development, and verification need to be addressed to 
enable adaptation to unknown situations, we focus on aspects unique to 
learning-based systems, \ie concerning the learning process itself, as 
the main driver to achieve adaptation to unknown situations. We focus on three main categories: (i) autonomously building environment representations, (ii) retainment, reuse, and modification of 
available knowledge required for lifelong learning, and (iii) autonomous 
goal adaptation.

\subsection{Environment observation representation}
Software systems are equipped with sensors and monitors 
which enable them to observe their external and internal environment, together with the structures required to represent relevant 
environment information. In case of unknown situations in the 
environment, the system might not have the capabilities to sense the 
new state space, to represent it, or to classify it as relevant for 
the fulfillment of its goals. Early work in this area addresses 
dynamic adaptations of state-space representation in existing learning 
processes~\cite{gueriau19}, however, currently, the adaptation refers 
only to the granularity of the sensed information with similar meaning 
(\eg cold vs -5 degrees). Together with the state-space granularity 
representation, the system first needs to identify the relevant 
dimensions of sensed information (\eg identifying if weather 
information is relevant at all), and then which data from each dimension is 
relevant (\eg temperature, humidity, or atmospheric pressure). Only 
then it would be possible for learning processes to incorporate newly 
sensed information from unknown situations. 

\subsection{Lifelong learning and adaptation evolution}
After successfully sensing and representing new situations, 
self-adaptive systems need to generate brand new behaviour to adapt to 
such situations. Using learning processes to learn this from scratch 
at run time could be detrimental to system performance. Various 
approaches are being explored to derive new behaviour from previously 
experienced situations but advancements will need to be made in 
multiple research sub-areas depending on the level of similarity of the 
new situations. In simpler cases, new behavior might be a 
composition of existing adaptations~\cite{cardozo20comina} (which 
requires managing potential conflicts between adaptations' behaviour), 
or might only require parameter evolution to adapt previously seen 
situations. A repository of behaviour classified by environment 
conditions can be maintained, using similarity metrics to identify the 
closest matches, and initiate a new learning process bootstrapped with 
"close-enough" existing behaviour for online 
fine-tuning~\cite{marinescu17}. However, scaling this to the execution 
of a self-adaptive system with multiple dissimilar situations, 
requires complex continual (lifelong) learning 
techniques~\cite{khetarpal20}.  
Existing open issues to address include identifying similarities 
between situations, transfer learning between tasks, and limiting 
forgetting of previous useful behaviour as numerous new situations arise.

\subsection{Goal adaptation and fulfillment}
Environment changes may also give rise to situations in which one or 
more of the system goals cannot be met successfully, regardless of its 
capabilities and knowledge. Similarly, new observations might require 
modifications or additions to the existing system goals. There are a 
number of considerations here that need to be addressed. First, the system 
needs to identify that it is not able to attain the goal. 
Second, the system must identify whether the goal is partially 
obtainable from explored states "close enough" to the goal, and 
whether partial fulfilment is preferable to system failure. Third, 
the system must identify suitable substitute 
goals, and proceed to learn to fulfil them, or interface with a human to 
propose newly defined goals. Flexibility of goals can be 
expressed within goal specification, \ie whether it is permitted for 
the system to modify a given goal. Online goal adaptation, as well as 
generation of the associated learning process to meet the adapted goal 
is likely to require significant research progress, given that even 
dynamic specification of individual parts of the learning process 
(\eg states, as discussed earlier) is in its early stages. However, 
progress is already being made on agents autonomously generating their 
own goals using generative adversarial networks~\cite{florensa18} 
in order to identify a range of tasks/skills that are possible to 
achieve. In the context of self-adaptive systems, such goals can be 
initially used as suggestions to human in the loop (\eg "I cannot 
currently reach the goal, but these are the states I can reach"), 
or for fully autonomous goal adaptation where such feedback can come 
from interaction with environment.


\section{Final Remarks}
\label{sec:conclusion}

Special consideration should be given to how to incorporate ethical 
rules into self-adaptive systems, if we equip them with the technical 
capabilities to fully operate, adapt and even modify their own goals 
without the human in the loop. Similarly to having to generate new 
behaviour to adapt to new situations, adapting to unknown situations 
might also involve making ethical choices that were not foreseen or 
that have not been previously encountered by the system, and therefore 
the system does not have human guidance on how to proceed. Efforts on 
enabling self-adaptation to unknown situations should progress 
hand-in-hand with efforts on developing a set of ethical principles 
of self-adaptive systems~\cite{weyns20}. Addressing research and 
technical challenges of encoding ethical aspects into self-adaptive 
systems should take priority before equipping them with capabilities 
to fully guide their own performance in all circumstances. 



\section*{Acknowledgments}
This publication is supported in part by a research grant from Science Foundation Ireland (SFI) under Grant Number 16/SP/3804.


\printbibliography

\end{document}

\endinput